%% file: main.tex
\documentclass[10pt,twocolumn,letterpaper]{article}

\usepackage[pagenumbers]{cvpr} 

\input{preamble}

\definecolor{cvprblue}{rgb}{0.21,0.49,0.74}
\usepackage[pagebackref,breaklinks,colorlinks,allcolors=cvprblue]{hyperref}

\def\confName{CVPR}
\def\confYear{2027}

\title{Wear and Go: Calibration Free Sparse Inertial Motion Capture\\via Online Sensor Setup Estimation}

\author{Yuanshuo Tan\\
Shanghai Jiao Tong University\\
{\tt\small tanyuanshuo@sjtu.edu.cn}
\and
Licheng Zhong\\
National University of Singapore\\
{\tt\small zlicheng@u.nus.edu}
}

\begin{document}
\twocolumn[{%
        \renewcommand\twocolumn[1][]{#1}%
        \maketitle
        \vspace{-1.5em}
        \begin{center}
  \includegraphics[width=\linewidth]{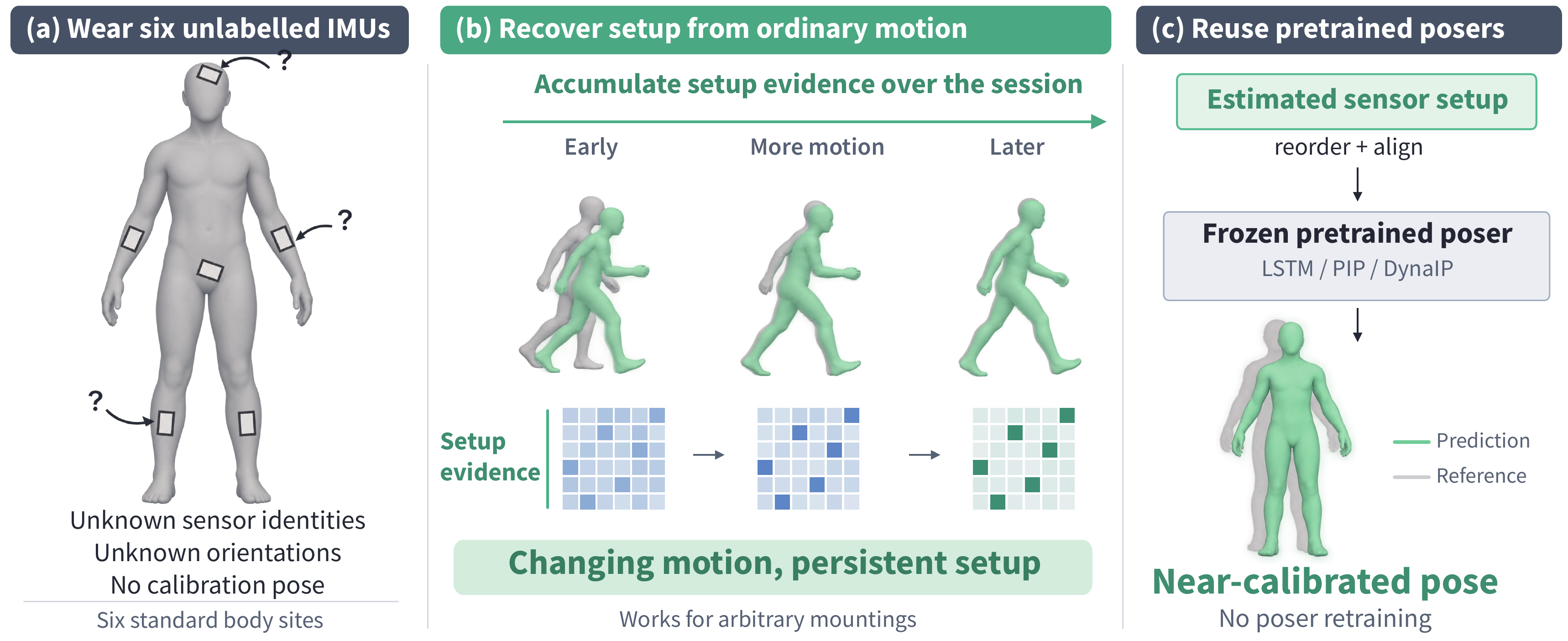}\vspace{-0.5em}
  \captionof{figure}{\textbf{\method{} enables calibration free inertial motion capture.}
Users attach six interchangeable IMUs to prescribed body sites
without specifying sensor identities or arbitrary mounting orientations
and begin moving without a calibration pose.
\method{} accumulates evidence from ordinary motion to recover
the persistent sensor setup, progressively aligning the measurements
for accurate reconstruction with frozen pose networks. Project page: \url{https://tanyuanshuo.github.io/wearandgo_page/}
}
  \label{fig:teaser}
        \end{center}
        \vspace{0.6em}
    }]

\input{sec/0_abstract}
\input{sec/1_intro}
\input{sec/fig_pipeline}
\input{sec/2_related}
\input{sec/3_method}
\input{sec/tab_main}
\input{sec/tab_pip}
\input{sec/4_experiments}

\input{sec/5_conclusion}
\input{sec/6_llm_usage}
{
    \small
    \bibliographystyle{ieeenat_fullname}
    \bibliography{main}
}
\appendix
\input{sec/X_suppl}

\end{document}

%% file: preamble.tex
\usepackage{cuted} 

\usepackage{currfile} 

\usepackage{caption} 

\newcommand{\qheading}[1]{\noindent\mbox{\textbf{#1}\;}}

\newcommand{\method}{MOSAIC}

\usepackage{booktabs}
\usepackage{amsmath,amssymb,bm}
\usepackage{tikz}
\usetikzlibrary{positioning,arrows.meta}

%% file: sec/0_abstract.tex
\begin{abstract}
Sparse inertial motion capture typically assumes that each sensor's anatomical identity and mounting orientation are known.
We present MOSAIC (Motion-based Sensor Assignment and
Inertial Calibration), a framework that recovers this setup from ordinary motion without a calibration pose.
Our central insight is that body pose changes continuously while the sensor setup persists, allowing movements that are individually ambiguous to provide complementary evidence over time.
A setup network processes unordered sensor streams with temporal and cross-sensor attention, then accumulates assignment and rotation evidence into a session-level estimate.
The recovered setup canonicalizes measurements for frozen pretrained pose estimators, separating sensor configuration from motion reconstruction.
For sensors worn at prescribed body sites, evaluation on real inertial recordings with randomized setups shows near-calibrated pose accuracy for bounded offsets and substantial benefits from accumulation under arbitrary orientations.
Transfer across pose networks supports setup recovery as a reusable interface between interchangeable sensors and pretrained pose models.
\end{abstract}
\vspace{-3mm}

%% file: sec/1_intro.tex
\section{Introduction}\label{sec:intro}
Sparse inertial motion capture reconstructs full-body movement from a few inertial measurement units (IMUs), supporting animation and motion analysis beyond camera-equipped capture spaces~\cite{vonmarcard2017sip,huang2018dip,mollyn2023imuposer}.
Modern systems exploit learned motion priors, kinematic structure and physical constraints to infer joints that are not directly observed~\cite{yi2021transpose,yi2022pip,zhang2024dynaip}.
Their inputs, however, must be associated with the correct body segments and expressed in the expected coordinate frames.
Assigning devices and performing a calibration pose add friction every time the sensors are worn.
Can a user attach interchangeable sensors at the prescribed sites and let ordinary movement recover the setup?

This requires estimating two coupled unknowns: a discrete sensor-to-segment assignment and a continuous mounting rotation for each device.
Dynamic calibration can estimate alignment during movement when segment identities are known~\cite{zuo2025tic}. Joint assignment and alignment have also been studied for lower-body walking~\cite{zimmermann2018}.
For sparse full-body capture during varied activity, the central difficulty is that setup information arrives unevenly.
A short window may distinguish the moving limb but leave its axial rotation weakly constrained.
Moreover, the forearms, shanks, head and pelvis do not form a directly instrumented chain, limiting the use of adjacent-joint constraints.

Our key observation is that \emph{setup is persistent even when pose is changing}.
Once the devices are attached, different movements provide repeated, complementary observations of the same assignment and mounting rotations.
We propose \method{} (Motion-based Sensor Assignment and Inertial Calibration), which exploits this persistence by extracting setup evidence from unordered sensor streams within each motion window and accumulating it over time (\cref{fig:teaser,fig:pipeline}).
Temporal attention captures each stream's motion and cross-sensor attention relates it to the rest of the body.
Accumulated assignment scores enforce a one-to-one anatomical mapping, and accumulated matrix evidence yields a mounting rotation for each sensor.
The estimate then reorders and aligns the measurements for a frozen pose network.
This separation lets setup evidence span a session while the pose estimator follows the current motion. 

The contributions of this paper are summarized as follows:
\begin{itemize}
  \item An online framework that jointly recovers the identities and mounting rotations of six interchangeable IMUs at prescribed sites, without a calibration pose.
  \item A sensor-order-equivariant setup network with persistent evidence accumulation and explicit canonicalization for frozen pose estimators.
  \item An analysis of how motion content and accumulated history determine setup recovery and downstream pose accuracy, including transfer across pretrained pose models.
\end{itemize}

%% file: sec/fig_pipeline.tex
\begin{figure*}[t]
  \centering
  \includegraphics[width=\linewidth]{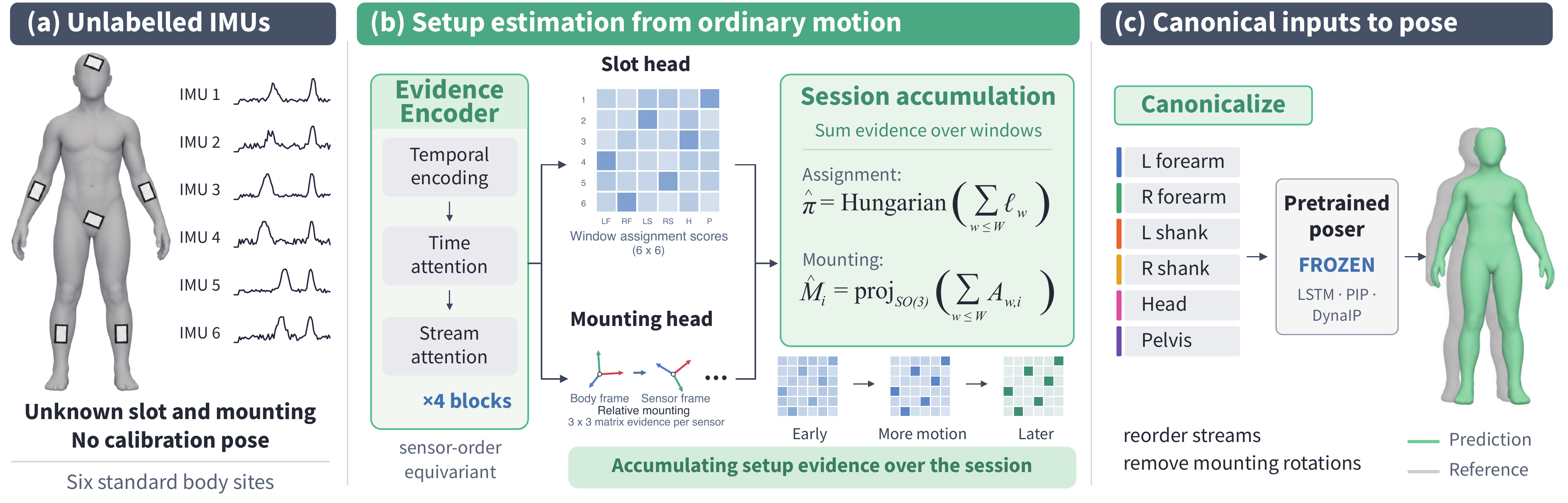}
  \caption{\textbf{Overview of \method{}.}
(a) The input consists of six IMU streams with unknown anatomical assignments and mounting rotations.
(b) A shared encoder alternates temporal and cross-sensor attention to extract window-level assignment scores and mounting evidence. These outputs accumulate across the session. Hungarian matching and projection onto $SO(3)$ recover the assignment and mounting rotations, respectively.
(c) The estimated setup canonicalizes the measurements for a frozen pose network, such as LSTM, PIP or DynaIP~\cite{yi2022pip,zhang2024dynaip}.
Green and gray meshes denote predicted and reference poses,
respectively.}
  \label{fig:pipeline}
\end{figure*}

%% file: sec/2_related.tex
\section{Related work}\label{sec:related}
\qheading{Sparse inertial pose estimation.}
SIP~\cite{vonmarcard2017sip} fits a body model to six IMUs, and DIP~\cite{huang2018dip} learns a motion prior for real-time reconstruction.
TransPose~\cite{yi2021transpose} introduces intermediate kinematic predictions and PIP~\cite{yi2022pip} adds physical constraints. DynaIP~\cite{zhang2024dynaip} learns velocity and part-specific dynamics.
DiffusionPoser~\cite{vanwouwe2024dp} supports varying sensor configurations and missing measurements through autoregressive diffusion.
These methods address reconstruction from sparse observations.
Our method supplies their required anatomical input convention by estimating the sensor setup before pose inference.

\smallskip
\qheading{Calibration, assignment and alignment.}
Static, functional and anatomical calibration use prescribed postures, joint movements or landmarks to identify sensor-to-segment transformations~\cite{favre2009,picerno2008,pacher2020review}.
TIC~\cite{zuo2025tic} estimates mounting offsets and sensor-specific heading drift during ongoing motion, given known segment identities.
Location recognition from walking~\cite{kunze2005where} and learned lower-body assignment and alignment~\cite{zimmermann2018} establish that motion can also identify where a sensor is worn.
JointTracker-based calibration~\cite{taetz2026}, joint-axis methods~\cite{seel2014,laidig2017} and RING~\cite{bachhuber2024ring} exploit kinematic-chain structure to recover alignment or relative orientation.
Our setting combines unknown sensor identities and mounting
rotations at six non-adjacent full-body sites.
We learn relationships across all sensor streams and accumulate
complementary evidence from natural motion to recover a persistent
setup for downstream pose estimation, without requiring a
prescribed calibration pose or motion sequence.

\smallskip
\qheading{Placement robustness and flexible sensing.}
Orientation- and displacement-tolerant activity recognition reduces sensitivity to how devices are worn~\cite{yurtman2017orient,banos2014displacement}.
For pose estimation, MODA~\cite{wu2025moda} augments training with sensor rotation and sliding, and Loose Inertial Poser~\cite{zuo2024lip} models garment-induced secondary motion.
Consumer-device systems~\cite{mollyn2023imuposer,xu2024mobileposer} and IMUCoCo~\cite{zhou2025imucoco} accommodate flexible sensing configurations with device-location information.
ArmPoser~\cite{dev2026armposer} handles smartwatch wear configurations and distinguishes forearm placement and crown orientation.
We explicitly recover the six-device assignment and mounting rotations, preserving the input interface of an existing poser.
Our invariant-poser baseline tests the alternative of learning mounting robustness directly in the pose network.

%% file: sec/3_method.tex
\section{Method}\label{sec:method}
\method{} separates persistent sensor setup from changing
body pose (\cref{fig:pipeline}).
A setup estimation network extracts sensor-to-segment
assignment and mounting-rotation evidence from short windows
of six unordered IMU streams.
This evidence is accumulated across windows to maintain
a session-level setup estimate, which reorders and aligns
incoming measurements for a frozen pose network.

\subsection{Setup model and inputs}\label{sec:model}
One IMU is attached to each forearm, each shank, the head and the pelvis, following the standard six-sensor layout~\cite{huang2018dip,yi2021transpose,yi2022pip}.
Anatomical identities and mounting orientations are unknown.
Stream indices track the same devices over time. 
The sensors provide orientations in a common world frame $\mathcal{W}$ with shared heading, as with magnetometer-aided fusion.
We define segment frames using SMPL~\cite{loper2015}.


Let $\mathcal{W}$ denote the common world frame,
$\mathcal{S}_i$ the local frame of sensor $i$, and
$\mathcal{B}_k$ the frame of body segment $k$.
The IMU observations comprise the sensor orientation
$\mathbf{O}^{IMU}_i=\mathbf{R}_{\mathcal{W}\mathcal{S}_i}$,
obtained from orientation fusion, and its gravity-removed
acceleration $\mathbf{a}^{IMU}_i$, expressed in $\mathcal{W}$.
The unknown assignment $\pi(i)$ identifies the segment
carrying sensor $i$, while
$\mathbf{M}_i=\mathbf{R}_{\mathcal{B}_{\pi(i)}\mathcal{S}_i}$
is its unknown sensor-to-segment mounting rotation.
Writing $\mathbf{O}_{\pi(i)}=\mathbf{R}_{\mathcal{W}\mathcal{B}_{\pi(i)}}$
for the segment orientation and $\mathbf{a}_k$ for the
world-frame gravity-removed acceleration at its attachment
site, we have
\begin{equation}
  \mathbf{O}^{IMU}_i=\mathbf{O}_{\pi(i)}\mathbf{M}_i,\qquad
  \mathbf{a}^{IMU}_i=\mathbf{a}_{\pi(i)}.
  \label{eq:setup}
\end{equation}
The setup $(\pi,\{\mathbf{M}_i\})$ stays fixed during rigid attachment.
We study bounded mounting offsets, $\angle\mathbf{M}_i\le\theta_{\max}$, and arbitrary orientations, $\mathbf{M}_i\in SO(3)$.
Changing the mounting orientation at a fixed site leaves world-frame acceleration unchanged; recovering $\mathbf{M}_i$ therefore requires relating orientation to body motion.
Across non-adjacent segments, these relationships are learned from whole-body movement.

Both mounting regimes directly use these standard sparse-IMU inputs,
flattening each sensor-to-world rotation matrix.
Training randomizes assignment, mounting and common heading, holding the same setup across windows of each sample.

\subsection{Window-level setup evidence}\label{sec:setupnet}
A shared linear embedding and strided temporal convolution encode each stream.
The backbone alternates temporal self-attention within each stream and cross-sensor self-attention at each time step.
Shared weights and the absence of sensor-identity embeddings make the network equivariant to stream order~\cite{zaheer2017deepsets,lee2019settransformer}.
After time pooling, another cross-sensor layer predicts slot log-probabilities $\boldsymbol\ell_w\in\mathbb{R}^{6\times6}$ and mounting evidence $\mathbf{A}_{w,i}\in\mathbb{R}^{3\times3}$ for window $w$.

For bounded offsets, the mounting head predicts an unconstrained matrix $\mathbf{A}_{w,i}$.
For arbitrary orientations, it predicts a rotation $\mathbf{R}_{w,i}$ using the 6D representation~\cite{zhou2019rotation} and a positive confidence $c_{w,i}=\mathrm{softplus}(\cdot)$, giving $\mathbf{A}_{w,i}=c_{w,i}\mathbf{R}_{w,i}$.
Both heads produce additive evidence in the same sensor-to-segment convention; the arbitrary head explicitly separates the rotation estimate from its weight.
The network predicts additive evidence before enforcing the final assignment and mounting estimates, allowing windows to support a common setup despite differing predictions.

\subsection{Persistent setup estimation}\label{sec:accum}
At each update, \method{} reads a trailing motion window from the current recording.
After $W$ updates, we recover the setup as
\begin{equation}
  \begin{aligned}
    \hat\pi &= \mathrm{Hungarian}\Big(\sum_{w\le W}\boldsymbol{\ell}_w\Big),\\
    \hat{\mathbf{M}}_i &= \mathrm{proj}_{SO(3)}\Big(\sum_{w\le W}\mathbf{A}_{w,i}\Big).
  \end{aligned}
  \label{eq:accum}
\end{equation}
Each new window adds its assignment scores and mounting evidence
to running sums maintained for each device. Hungarian matching maximizes the summed assignment score subject to one sensor per slot.
Given the singular value decomposition
$\mathbf{B}=\mathbf{U}\mathbf{S}\mathbf{V}^\top$,
its projection onto $SO(3)$ is
$\mathbf{U}\,\mathrm{diag}(1,1,\det(\mathbf{U}\mathbf{V}^\top))
\mathbf{V}^\top$.
Thus the session retains fixed-size sums of assignment and mounting evidence, regardless of duration.
Pooling takes place separately for each persistent device stream, before the estimated permutation is applied to pose inputs.
This prevents a transient assignment change from redirecting one device's mounting history to another.
Matrix Fisher distributions motivate this representation: their natural parameters add under products of densities, and projection recovers the mode~\cite{mohlin2020matrixfisher}.
The sliding windows overlap, so the sum is trained as a pooling operator over correlated observations; it does not assume independent evidence.

For the arbitrary-mounting head, the projected sum is also the confidence-weighted chordal rotation mean:
\begin{equation}
  \hat{\mathbf{M}}_i=\underset{\mathbf{R}\in SO(3)}{\arg\min}
  \sum_{w\le W}c_{w,i}\|\mathbf{R}-\mathbf{R}_{w,i}\|_F^2.
  \label{eq:chordal}
\end{equation}
Expanding the squared distance leaves a trace maximization against the summed evidence, which is solved by the projection in \cref{eq:accum}.
The bounded head instead learns matrix evidence directly, allowing its entries to encode how a window contributes to the accumulated rotation.
Both forms share the same inference rule and are supervised through the resulting setup.

Accumulation training samples $K$ windows from one sequence under one setup and supervises every prefix:
\begin{multline}
  \mathcal{L}=\frac1K\sum_{k=1}^{K}\Big[\mathrm{CE}\Big(\sum_{w\le k}\boldsymbol{\ell}_w,\pi\Big)\\
  +\frac16\sum_i\big\|\mathrm{proj}_{SO(3)}\big(\textstyle\sum_{w\le k}\mathbf{A}_{w,i}\big)-\mathbf{M}_i\big\|_F^2\Big].
  \label{eq:loss}
\end{multline}
For arbitrary orientations, accumulation training replaces
$\mathrm{proj}_{SO(3)}(\mathbf{B}_{k,i})$ in the mounting-loss
term of \cref{eq:loss} with
$\sqrt{3}\,\mathbf{B}_{k,i}/
\max(\|\mathbf{B}_{k,i}\|_F,\epsilon)$,
where $\mathbf{B}_{k,i}=\sum_{w\leq k}\mathbf{A}_{w,i}$
and $\epsilon>0$ ensures numerical stability.
This surrogate avoids unstable SVD gradients when rotations agree.
Inference always uses \cref{eq:accum}.
Prefix supervision makes the evidence useful both on its own and in combination with other windows.
The mounting loss acts on the pooled estimate, so the evidence head learns how a window should contribute to recovery of the shared setup.

At a cold start, the first motion block supplies the initial estimate and is reconstructed after this brief warm-up.
Subsequent blocks use only preceding observations.
Setup evidence persists across recordings within a session.
For re-mounting, a parallel state tracks recent evidence with exponential forgetting.
Once sufficient history has accumulated, a large angular discrepancy between recent and session estimates replaces the affected stream's mounting sum with its recent evidence.
This adapts mounting estimates while preserving the assignment and other sensors' histories.

\subsection{Canonicalization and pose reconstruction}\label{sec:canon}
Let $\mathbf{H}$ be the fixed yaw rotation that aligns the initial corrected pelvis with the poser's reference forward direction~\cite{yi2021transpose}.
We construct the canonical inputs as
\begin{equation}
  \begin{aligned}
    \tilde{\mathbf{O}}_{\hat\pi(i)} &= \mathbf{H}\mathbf{O}^{IMU}_i\hat{\mathbf{M}}_i^\top,\\
    \tilde{\mathbf{a}}_{\hat\pi(i)} &= \mathbf{H}\mathbf{a}^{IMU}_i.
  \end{aligned}
  \label{eq:canonical}
\end{equation}
The mounting correction acts in sensor coordinates, while the heading rotation acts in world coordinates and is shared across the recording.
We compute $\mathbf{H}$ from the measured initial orientation and estimated pelvis assignment and mounting, using the warm-up estimate at a cold start. No ground truth is used.
With the true assignment and mountings, \cref{eq:canonical} inverts the setup transformation exactly, up to the chosen common heading.
The canonicalized measurements feed a frozen pose network using its existing input preprocessing.
Its recurrent state and heading alignment reset at each recording, independently of the persistent setup state.

%% file: sec/tab_main.tex
\begin{table*}[t]
\centering
\small
\setlength{\tabcolsep}{2pt}
\caption{
\textbf{Pose accuracy under unknown setup.}
Results are averaged over recordings.
Per take resets setup evidence at each recording;
per session retains it across recordings.
$^\dagger$True sensor-to-segment assignments are available.
The calibrated-input reference uses dataset-provided calibrated
measurements without additional setup transformations.
Lower is better for all metrics.
}
\label{tab:main}

\begin{tabular*}{\textwidth}{@{\extracolsep{\fill}}l*{10}{c}@{}}
\toprule
& \multicolumn{5}{c}{DIP-IMU}
& \multicolumn{5}{c}{TotalCapture} \\
\cmidrule(lr){2-6}
\cmidrule(l){7-11}
Method
& SIP ($^\circ$) & Ang ($^\circ$) & Pos (cm)
& Mesh (cm) & Jitter (km/s$^3$)
& SIP ($^\circ$) & Ang ($^\circ$) & Pos (cm)
& Mesh (cm) & Jitter (km/s$^3$) \\
\midrule
\textcolor{gray}{Calibrated input$^\dagger$}
& \textcolor{gray}{16.5} & \textcolor{gray}{9.7}
& \textcolor{gray}{5.4} & \textcolor{gray}{6.4}
& \textcolor{gray}{2.18}
& \textcolor{gray}{12.7} & \textcolor{gray}{11.7}
& \textcolor{gray}{5.2} & \textcolor{gray}{6.0}
& \textcolor{gray}{3.85} \\
\midrule
\multicolumn{11}{@{}l}{
\emph{Bounded mounting ($\theta_{\max}=30^\circ$)}
} \\
No estimate
& 67.2 & 63.6 & 28.5 & 34.4 & 3.72
& 62.0 & 59.7 & 27.0 & 33.2 & 5.35 \\
No calibration$^\dagger$
& 28.6 & 23.0 & 10.8 & 12.7 & 2.50
& 23.1 & 22.4 & 10.5 & 12.3 & 4.26 \\
TIC~\cite{zuo2025tic}, retrained$^\dagger$
& 26.7 & 21.2 & 9.9 & 11.6 & 2.41
& 17.8 & 16.8 & 7.8 & 9.0 & 4.00 \\
Invariant poser, per take
& \textbf{16.2} & 14.2 & 6.1 & 7.3 & 3.23
& 19.3 & 16.7 & 7.1 & 8.4 & 5.53 \\
Invariant poser, per session
& \textbf{16.2} & 14.1 & 6.1 & 7.3 & 3.21
& 19.3 & 16.7 & 7.1 & 8.4 & 5.53 \\
Ours, per take
& 17.0 & 11.9 & 5.7 & 6.8 & \textbf{2.18}
& 14.0 & 13.2 & 5.8 & 6.8 & \textbf{3.89} \\
Ours, per session
& 16.8 & \textbf{11.2} & \textbf{5.6}
& \textbf{6.7} & \textbf{2.18}
& \textbf{13.6} & \textbf{12.9} & \textbf{5.6}
& \textbf{6.5} & \textbf{3.89} \\
\midrule
\multicolumn{11}{@{}l}{
\emph{Arbitrary mounting ($SO(3)$)}
} \\
No estimate
& 71.9 & 72.4 & 31.1 & 38.1 & 3.85
& 74.8 & 72.6 & 32.6 & 39.8 & 6.02 \\
No calibration$^\dagger$
& 67.8 & 69.7 & 29.5 & 36.4 & 3.69
& 71.2 & 71.0 & 30.5 & 37.3 & 5.67 \\
TIC~\cite{zuo2025tic}, retrained$^\dagger$
& 60.7 & 61.4 & 25.7 & 31.6 & 3.47
& 39.2 & 38.9 & 17.2 & 20.5 & 4.45 \\
Invariant poser, per take
& 20.3 & 22.0 & 8.3 & 10.4 & 3.46
& 27.1 & 26.9 & 11.1 & 13.9 & 4.59 \\
Invariant poser, per session
& 20.0 & 21.8 & 8.2 & 10.3 & 3.44
& 26.9 & 26.7 & 11.0 & 13.7 & 4.57 \\
Ours, per take
& 25.0 & 20.3 & 8.4 & 10.5 & 2.45
& 18.3 & 17.9 & 7.8 & 9.4 & 4.08 \\
Ours, per session
& \textbf{19.2} & \textbf{13.7} & \textbf{6.4}
& \textbf{7.7} & \textbf{2.25}
& \textbf{14.2} & \textbf{13.4} & \textbf{5.9}
& \textbf{6.9} & \textbf{3.96} \\
\bottomrule
\end{tabular*}
\end{table*}

%% file: sec/tab_pip.tex
\begin{table}[t]
\centering
\small
\setlength{\tabcolsep}{2.5pt}
\caption{\textbf{Compatibility with advanced pose networks.}
PIP uses pose predictions before physics-based optimization;
DynaIP follows the default poser's training data and schedule.}
\label{tab:pip}
\begin{tabular}{@{}llrrr@{}}
\toprule
Network & Input & SIP ($^\circ$) & Ang ($^\circ$) & Pos (cm) \\
\midrule
\multicolumn{5}{c}{DIP-IMU} \\
\midrule
PIP
 & Calibrated input
 & 15.4 & 8.5 & 4.8 \\
 & Ours ($\theta_{\max}=30^\circ$)
 & 15.8 & 10.3 & 5.0 \\
 & Ours (arbitrary)
 & 17.5 & 12.6 & 5.6 \\
\addlinespace
DynaIP
 & Calibrated input
 & 14.9 & 7.2 & 4.7 \\
 & Ours ($\theta_{\max}=30^\circ$)
 & 15.1 & 9.8 & 5.1 \\
 & Ours (arbitrary)
 & 16.2 & 12.2 & 5.8 \\
\midrule
\multicolumn{5}{c}{TotalCapture} \\
\midrule
PIP
 & Calibrated input
 & 12.8 & 11.8 & 5.2 \\
 & Ours ($\theta_{\max}=30^\circ$)
 & 14.1 & 13.1 & 5.7 \\
 & Ours (arbitrary)
 & 14.8 & 13.7 & 6.1 \\
\addlinespace
DynaIP
 & Calibrated input
 & 13.0 & 11.8 & 5.1 \\
 & Ours ($\theta_{\max}=30^\circ$)
 & 13.6 & 12.9 & 5.5 \\
 & Ours (arbitrary)
 & 14.1 & 13.6 & 5.9 \\
\bottomrule
\end{tabular}
\end{table}

%% file: sec/4_experiments.tex
\section{Experiments}\label{sec:exp}

\subsection{Experimental setup}\label{sec:protocol}
\qheading{Data and protocol.}
Following prior work~\cite{yi2021transpose,yi2022pip},
we pretrain on synthetic IMU measurements from
AMASS~\cite{mahmood2019amass} and fine-tune on real IMU
recordings from DIP-IMU subjects 1--8~\cite{huang2018dip},
mixed with synthetic AMASS data.
We evaluate on real IMU recordings from the held-out
DIP-IMU subjects 9--10 and
TotalCapture~\cite{trumble2017totalcapture}.
To evaluate recovery from unknown sensor configurations, we transform calibrated recordings with random sensor permutations, a shared yaw rotation, and either bounded mounting offsets ($\angle\mathbf{M}_i\le\theta_{\max}$, with $\theta_{\max}=30^\circ$) or arbitrary orientations ($\mathbf{M}_i\in SO(3)$).
For each subject, we sample five sensor configurations and evaluate all methods on the same configurations. We compare two protocols: \emph{per-take}, which resets setup evidence for each recording, and \emph{per-session}, which accumulates evidence across a subject's recordings under a fixed configuration.

\smallskip
\qheading{Metrics.}
We evaluate pose reconstruction using five metrics commonly reported in sparse inertial motion capture~\cite{yi2021transpose,yi2022pip}:
1) \emph{SIP error}, the average global rotation discrepancy for the upper arms and thighs;
2) \emph{angular error (Ang)}, the corresponding average across all joints;
3) \emph{position error (Pos)}, the average distance between predicted and ground-truth joint positions relative to the root;
4) \emph{mesh error}, the average distance between corresponding SMPL vertices after root-position alignment; and
5) \emph{jitter}, the average magnitude of the third temporal derivative of predicted joint positions.
Setup recovery is evaluated using
6) \emph{recovery time}, the time until all assignments are correct and mean mounting error falls below a specified threshold; and 
7) \emph{mounting error}, the geodesic distance between estimated and true mounting rotations.


\smallskip
\qheading{Baselines.}
\emph{Calibrated input} supplies known sensor identities and mountings.
\emph{No estimate} uses unordered, uncorrected streams. \emph{No calibration} supplies true identities. 
\emph{TIC}~\cite{zuo2025tic} receives true sensor-to-segment
assignments and is retrained separately for bounded and
arbitrary mountings using our training data and matching
mounting distributions.
Its official online calibration procedure is applied
independently to each recording.
The \emph{invariant poser} replaces explicit mounting
correction with random-rotation augmentation, using \method{}'s estimated assignments
to reorder its inputs.

\smallskip
\qheading{Implementation.}
Main experiments use a frozen pretrained LSTM poser.
Compatibility tests use released PIP~\cite{yi2022pip}
weights and DynaIP~\cite{zhang2024dynaip} trained with
the same training data and schedule as our default poser.
Separate \method{} models are trained for bounded and arbitrary mountings.
The LSTM poser has three recurrent layers with 512 hidden units each, while \method{} uses four alternating attention blocks with 256-dimensional tokens.
\method{} updates once per second using up to 10\,s of motion, with a one-second cold-start warm-up.

\input{sec/fig_qualitative}
\input{sec/tab_setup}

\subsection{Pose accuracy and transfer}\label{sec:exp_main}\label{sec:exp_pip}
\qheading{Recovery of calibrated accuracy.}
Under bounded mountings, \method{}'s per-session gaps to calibrated
inputs are 1.1--1.5$^\circ$ in Ang and 0.3--1.0$^\circ$ in SIP
on both datasets (\cref{tab:main}), despite unknown identities and mountings.
Retrained TIC has a 5.1--11.5$^\circ$ Ang gap despite known assignments.
Thus, motion-based alignment lets a frozen poser approach calibrated
accuracy without a calibration pose.
Under arbitrary mountings, retaining evidence across recordings reduces
SIP by 23\% on DIP-IMU and 22\% on TotalCapture versus per-take recovery,
supporting persistent setup refinement from complementary motion.

\smallskip
\qheading{Qualitative results.}
\Cref{fig:qual} shows \method{} with the frozen LSTM poser.
In this session, motion evidence accumulates to resolve the unknown setup during
the first 10\,s; subsequent reconstructions closely resemble the
calibrated-input reference under both mounting regimes.

\smallskip
\qheading{Explicit versus implicit mounting compensation.}
The invariant poser learns mounting robustness through random-rotation
augmentation and uses the same estimated assignments, isolating implicit
compensation from explicit mounting estimation and correction.
Explicit correction improves most per-session pose metrics across both
datasets and regimes, consistently lowering angular, positional, and mesh
errors; arbitrary-mounting SIP on TotalCapture falls from 26.9$^\circ$
to 14.2$^\circ$.
This supports treating mountings as persistent physical parameters whose
correction brings frozen-poser inputs closer to calibrated training conditions.


\smallskip
\qheading{Compatibility with advanced pose networks.}
Beyond the default LSTM, frozen PIP~\cite{yi2022pip} and
DynaIP~\cite{zhang2024dynaip} receive corrected IMU inputs
without architectural changes (\cref{tab:pip}).
Under $\theta_{\max}=30^\circ$, SIP increases over calibrated inputs
by 0.4$^\circ$/1.2$^\circ$ for PIP and 0.1$^\circ$/0.6$^\circ$
for DynaIP on DIP-IMU/TotalCapture, respectively.
Under arbitrary mountings, DynaIP stays within 1.3$^\circ$
of calibrated SIP accuracy on both datasets.

\input{sec/tab_accum}
\input{sec/tab_oracle}

\input{sec/fig_time}

\subsection{Online setup recovery}\label{sec:exp_time}
\Cref{fig:time,tab:setup} show how setup recovery improves
as motion evidence accumulates.
For bounded mountings, all six sensor-to-segment assignments
are recovered within 2\,s in every evaluated session
on both datasets and remain correct thereafter.
The median time to correct assignment and mean mounting
error below $15^\circ$ is 1\,s.
At 10\,s, mean mounting errors are 6.2$^\circ$ on DIP-IMU
and 7.8$^\circ$ on TotalCapture.
For arbitrary mountings, all assignments become correct
and remain correct within 8\,s on DIP-IMU and 4\,s
on TotalCapture.
The median times to the $15^\circ$ mounting-error criterion
with correct assignment are also 8\,s and 4\,s, respectively,
with all evaluated sessions reaching the threshold.
At 10\,s, mean mounting errors are 14.2$^\circ$ on DIP-IMU
and 12.1$^\circ$ on TotalCapture.
\Cref{fig:obs} further decomposes mounting errors
under arbitrary mountings into twist about each segment's longitudinal axis and swing of that axis.
These results show that mounting estimates continue
to improve as motion evidence accumulates,
after sensor assignments have been resolved.

\input{sec/fig_observability}
\input{sec/fig_live_demo}

\subsection{Ablations and analysis}\label{sec:exp_accum}\label{sec:exp_motion}
\qheading{What makes accumulation effective?}
\Cref{tab:accum} isolates window selection and evidence aggregation
with fixed networks and windows of up to 10\,s.
\emph{Last window} uses only the latest window's evidence;
\emph{non-overlapping} accumulates windows every 10\,s.
\emph{Rotation pooling} and our \emph{evidence pooling} accumulate
overlapping windows every second throughout the session.
The former projects each window's evidence onto $SO(3)$ before
summation; the latter sums predicted matrices directly.
Both sums are projected onto $SO(3)$.

For arbitrary mountings, history reduces SIP from 25.4$^\circ$ to
19.2$^\circ$ on DIP-IMU and from 20.0$^\circ$ to 14.2$^\circ$
on TotalCapture.
Non-overlapping windows retain most of this gain, supporting
complementary motion history; evidence pooling improves on
rotation pooling by a further 0.0--0.2$^\circ$.

\qheading{Which setup errors affect pose?}
\Cref{tab:oracle} replaces assignment, mounting, or both with ground truth.
True assignments barely change per-session accuracy, consistent with
all six assignments becoming correct within 8\,s on both datasets
under arbitrary mountings (\cref{tab:setup}).
With estimated assignments fixed, estimated rather than true mountings
increase SIP by only 0.3--2.4$^\circ$ across datasets and regimes,
preserving most of the calibrated pose accuracy.


\subsection{Robustness and runtime}\label{sec:exp_robust}
\qheading{Mounting range and re-attachment.}
When evaluated beyond its training range of
$\theta_{\max}=30^\circ$ at $\theta_{\max}=60^\circ$,
the bounded model's per-session SIP error increases
from 16.8$^\circ$ to 20.1$^\circ$ on DIP-IMU
and from 13.6$^\circ$ to 18.1$^\circ$ on TotalCapture.
Under arbitrary mountings, we simulate re-attachment by
replacing one sensor's mounting rotation with a new arbitrary
rotation before the middle recording.
The mounting-state reset mechanism described in
\cref{sec:accum} reduces SIP error from 24.4$^\circ$ to
23.2$^\circ$ on DIP-IMU and from 20.5$^\circ$ to
19.2$^\circ$ on TotalCapture, without affecting results
on sessions with unchanged mountings.

\smallskip
\qheading{Runtime.}
On an RTX 4070 Ti SUPER, \method{} takes 1.8--1.9\,ms per update.
The frozen LSTM poser takes 0.38\,ms per frame, including input
preprocessing and output rotation conversion.


\subsection{Live Demo}\label{sec:live_demo}

We evaluate \method{} in a live setting using six Xsens
IMUs worn on the forearms, shanks, head, and pelvis.
Before each trial, the six IMUs are randomly redistributed
among the prescribed body sites and reattached with
arbitrary mounting orientations.
All accumulated setup evidence is reset before each trial,
following the per-take protocol in \cref{sec:protocol}. 
\method{} processes the IMU streams at 60\,Hz and feeds
corrected measurements to the LSTM poser for online
SMPL reconstruction, with visualization at 30\,fps
(\cref{fig:live_demo}).
A RealSense camera records RGB video, from which
GVHMR~\cite{shen2024world} provides an offline visual
reference used only for quantitative evaluation.
Across three two-minute trials, all six
sensor-to-segment assignments are recovered within 7\,s
and remain correct through the end of each trial.
Against this visual reference, the mean SIP error, angular error, positional error, mesh error, and jitter are 22.4$^\circ$, 22.2$^\circ$,
8.7\,cm, 11.1\,cm, and 1.70\,km/s$^3$, respectively.
Further experimental details and results are provided
in the supplementary material.

%% file: sec/fig_qualitative.tex
\begin{figure*}[t]
  \centering
  \includegraphics[width=\linewidth]{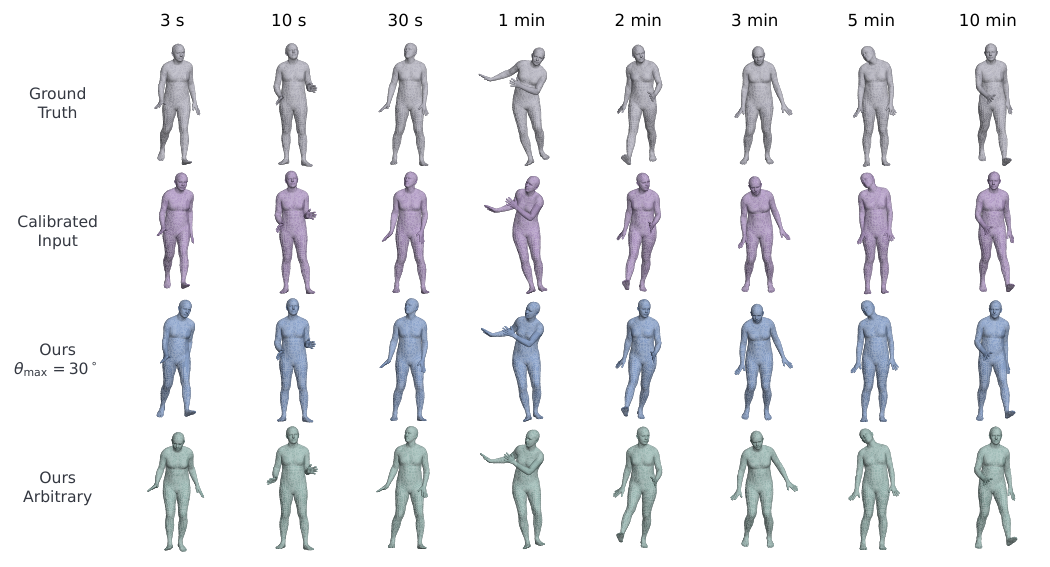}
  \caption{\textbf{Qualitative comparison of temporal pose reconstructions}
in a TotalCapture session. Rows show ground truth, the frozen
LSTM poser with calibrated input, and \method{} with the same
poser under bounded ($\theta_{\max}=30^\circ$) and arbitrary
mountings.}
  \label{fig:qual}
\end{figure*}

%% file: sec/tab_setup.tex
\begin{table}[t]
\centering\small
\setlength{\tabcolsep}{3pt}
\caption{\textbf{Online setup recovery.}
$t_{\mathrm{assign}}^{\max}$ is the maximum time across evaluated
sessions until all six sensor-to-segment assignments become
correct and remain correct thereafter.
$t_{15^\circ}$ is the median time to first achieve correct
assignment and mean mounting error below $15^\circ$;
all evaluated sessions reach this threshold.
Calib. error denotes the mounting rotation error at 10\,s,
averaged over sensors and sessions.}
\label{tab:setup}
\begin{tabular}{@{}llccc@{}}
\toprule
Mounting & Dataset
& $t_{\mathrm{assign}}^{\max}$ (s)
& $t_{15^\circ}$ (s)
& Calib. error ($^\circ$) \\
\midrule
$\theta_{\max}=30^\circ$ & DIP-IMU
& 2 & 1 & 6.2 \\
$\theta_{\max}=30^\circ$ & TotalCapture
& 2 & 1 & 7.8 \\
Arbitrary & DIP-IMU
& 8 & 8 & 14.2 \\
Arbitrary & TotalCapture
& 4 & 4 & 12.1 \\
\bottomrule
\end{tabular}
\end{table}

%% file: sec/tab_accum.tex
\begin{table}[t]
\centering
\small
\setlength{\tabcolsep}{3pt}
\caption{
\textbf{Ablation of temporal evidence accumulation.}
Per-session SIP error ($^\circ$)
with fixed networks and identical input recordings.
Best results within each mounting regime are bold.
}
\label{tab:accum}
\begin{tabular}{@{}lrr@{}}
\toprule
Strategy & DIP-IMU & TotalCapture \\
\midrule
\multicolumn{3}{@{}l}{
\emph{Bounded mounting ($\theta_{\max}=30^\circ$)}
} \\
Last window             & 18.0          & 14.8 \\
Non-overlapping         & \textbf{16.8} & 13.7 \\
Rotation pooling        & \textbf{16.8} & 13.7 \\
Evidence pooling (ours) & \textbf{16.8} & \textbf{13.6} \\
\midrule
\multicolumn{3}{@{}l}{\emph{Arbitrary mounting}} \\
Last window             & 25.4          & 20.0 \\
Non-overlapping         & 20.1          & 14.9 \\
Rotation pooling        & 19.4          & 14.3 \\
Evidence pooling (ours) & \textbf{19.2} & \textbf{14.2} \\
\bottomrule
\end{tabular}
\end{table}

%% file: sec/tab_oracle.tex
\begin{table}[t]
\centering
\small
\setlength{\tabcolsep}{3pt}
\caption{
\textbf{Ablation of setup errors.}
Per-session pose errors with ground-truth (GT)
assignment, mounting, or both.
}
\label{tab:oracle}
\begin{tabular}{@{}llrrrr@{}}
\toprule
& & \multicolumn{2}{c}{DIP-IMU}
  & \multicolumn{2}{c}{TotalCapture} \\
\cmidrule(lr){3-4}
\cmidrule(l){5-6}
Assignment & Mounting
& SIP ($^\circ$) & Ang ($^\circ$)
& SIP ($^\circ$) & Ang ($^\circ$) \\
\midrule
\multicolumn{6}{@{}l}{
\emph{Bounded mounting ($\theta_{\max}=30^\circ$)}
} \\
Estimated & Estimated & 16.8 & 11.2 & 13.6 & 12.9 \\
GT        & Estimated & 16.8 & 11.2 & 13.6 & 12.9 \\
Estimated & GT        & 16.5 &  9.7 & 12.7 & 11.7 \\
\midrule
\multicolumn{6}{@{}l}{\emph{Arbitrary mounting}} \\
Estimated & Estimated & 19.2 & 13.7 & 14.2 & 13.4 \\
GT        & Estimated & 19.2 & 13.7 & 14.2 & 13.3 \\
Estimated & GT        & 16.8 & 10.0 & 12.8 & 11.8 \\
\midrule
\multicolumn{6}{@{}l}{\textcolor{gray}{\emph{Calibrated reference}}} \\
\textcolor{gray}{GT}
& \textcolor{gray}{GT}
& \textcolor{gray}{16.5}
& \textcolor{gray}{9.7}
& \textcolor{gray}{12.7}
& \textcolor{gray}{11.7} \\
\bottomrule
\end{tabular}
\end{table}

%% file: sec/fig_time.tex
\begin{figure*}[t]
  \centering
  \includegraphics[width=\linewidth]{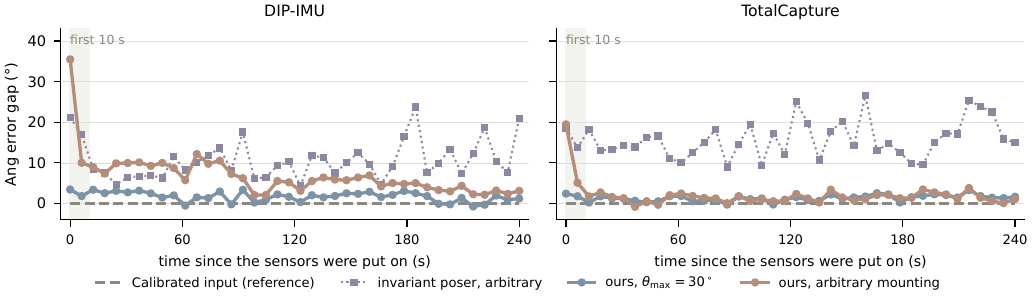}
  \caption{\textbf{Angular-error gap to the calibrated-input reference over time.}
Per-session differences in ground-truth-referenced angular error
are computed against calibrated inputs at matching frames and
averaged across subjects and sensor configurations at 40 uniformly
sampled times over the first four minutes.
The dashed zero line marks calibrated accuracy.}
  \label{fig:time}
\end{figure*}

%% file: sec/fig_observability.tex
\begin{figure*}[t]
  \centering
  \includegraphics[width=\linewidth]{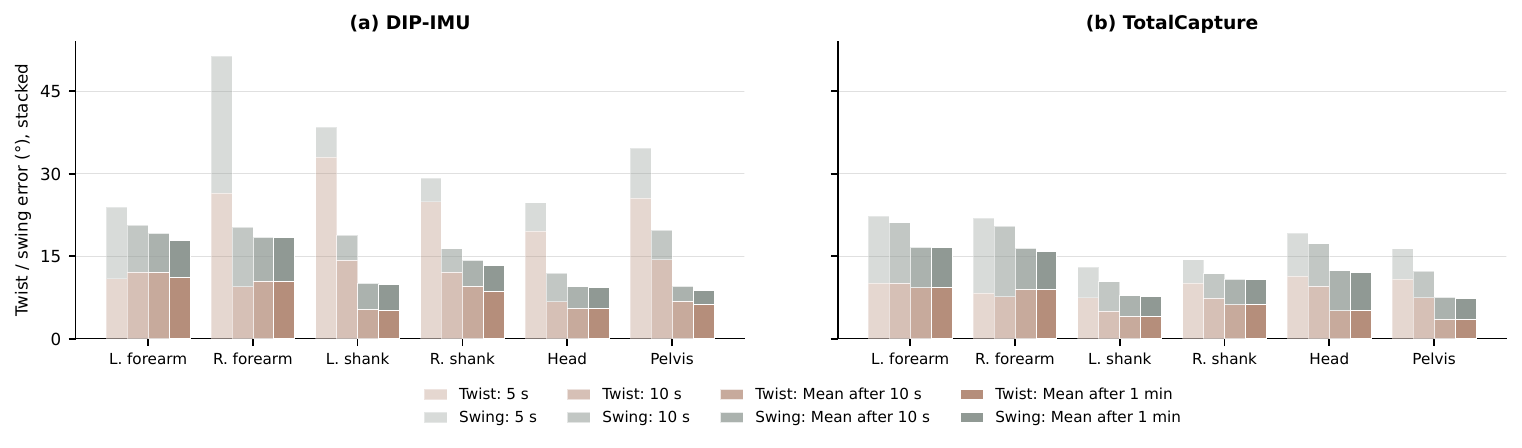}
  \caption{\textbf{Per-segment mounting recovery for arbitrary orientations, per session.}
Stacked twist/swing errors at 5\,s and 10\,s after session start,
and temporal means over the remainder after 10\,s and 1\,min.
Temporal means are computed within sessions before averaging across
subjects and sensor configurations.}
  \label{fig:obs}
\end{figure*}

%% file: sec/fig_live_demo.tex
\begin{figure*}[t]
  \centering
  \includegraphics[width=\linewidth]{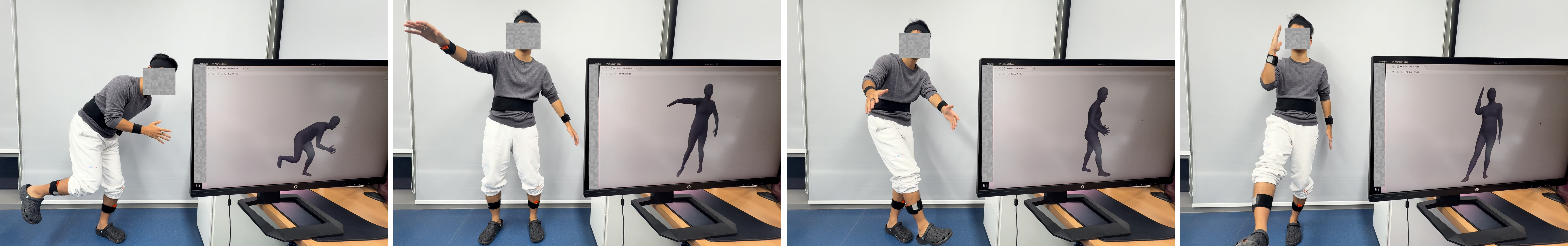}
  \caption{\textbf{Live demo.} Video frames and corresponding SMPL
  reconstructions using six IMUs with unknown sensor-to-segment
  assignments and arbitrary mounting orientations.}
  \label{fig:live_demo}
\end{figure*}

%% file: sec/5_conclusion.tex
\section{Conclusion}\label{sec:conclusion}
We presented \method{} for sparse inertial motion capture with unknown
sensor identities and mountings, without a calibration pose.
Accumulating complementary evidence from ordinary motion about the
persistent setup resolves single-window ambiguities and enables pretrained
pose networks to process unfamiliar configurations without retraining.
On real DIP-IMU and TotalCapture recordings, this yields near-calibrated
accuracy under bounded offsets, substantial gains from motion history
under arbitrary orientations, and transfer to PIP and DynaIP.
We further evaluate the framework on real-world IMU recordings that we collected under physical re-attachment, using vision-based pose estimates as a visual reference.

\smallskip
\qheading{Limitations.} We assume six IMUs at prescribed sites and constant
sensor-to-segment offsets. Resets accommodate mounting changes, but slippage
from loose straps or clothing remains unmodeled. Future work could address flexible layouts and time-varying sensor-to-segment alignment.

%% file: sec/6_llm_usage.tex
\section*{LLM Usage Statement}
Claude Code, ChatGPT, and Codex assisted with implementation and experiment scripts, analysis and presentation of experimental outputs, and manuscript drafting and revision.
All content has been reviewed by the authors, who take full responsibility for the manuscript and all results.

%% file: sec/X_suppl.tex
\clearpage
\twocolumn[{\raggedright\LARGE\bfseries Appendix\par\vspace{1em}}]

\section{Implementation details}
\label{sec:supp_impl}

\paragraph{Network architecture.}
Each sensor provides 12 input components: its world-frame
free acceleration divided by 30 and its flattened
sensor-to-world orientation matrix.
A shared linear projection maps these inputs to
256-dimensional tokens, followed by a temporal convolution
with kernel size 8, stride 4, and padding 2.
Sinusoidal temporal encodings are added to the resulting
sequence.
Four blocks alternate temporal self-attention within each
stream and cross-sensor self-attention at each time step.
Each layer uses pre-normalization, four attention heads and
a 1024-dimensional feed-forward layer.
After temporal averaging, one additional cross-sensor layer
feeds the assignment and mounting heads described in the
main paper. For arbitrary mountings, a separate linear layer maps
each sensor's feature after the final cross-sensor layer
to a scalar, followed by softplus to produce the
confidence weight.

\paragraph{Training \method{}.}
We uniformly sample window lengths between 60 and 600
frames at 60\,Hz and add Gaussian noise with standard
deviation 0.5\,m/s$^2$ to synthetic accelerations.
Single-window pretraining learns assignment and mounting
estimation from individual motion windows, without
accumulating evidence across windows.
For arbitrary mountings, this is followed by single-window
fine-tuning on mixed data.
Accumulation training groups four independently sampled
windows from the same recording under one shared sensor
configuration and supervises the setup estimate after
each successive evidence accumulation.
During this stage, the bounded model updates the entire
network, whereas the arbitrary model updates only its
confidence head.
Batch size counts individual windows in single-window
training and four-window groups in accumulation training.
\Cref{tab:supp_train} lists the training schedule.
All stages use AdamW with zero weight decay.
Mixed-data stages sample equal numbers of AMASS~\cite{mahmood2019amass} and
DIP-IMU~\cite{huang2018dip} windows.
All stages use their final checkpoints.

\begin{table*}[t]
\centering
\small
\setlength{\tabcolsep}{6pt}
\caption{\textbf{Training schedule for \method{}.}
Mixed data combines AMASS and DIP-IMU subjects 1--8
in equal proportions.}
\label{tab:supp_train}
\begin{tabular}{@{}lllrrc@{}}
\toprule
Mounting & Stage & Data
& Iterations & Batch size & Learning rate \\
\midrule
$\theta_{\max}=30^\circ$
& Single-window pretraining & AMASS
& 30,000 & 64 & $3\times10^{-4}$ \\
& Accumulation training & Mixed
& 8,000 & 16 & $10^{-4}$ \\
\addlinespace
Arbitrary
& Single-window pretraining & AMASS
& 30,000 & 64 & $3\times10^{-4}$ \\
& Single-window fine-tuning & Mixed
& 6,000 & 64 & $10^{-4}$ \\
& Accumulation training (confidence only) & Mixed
& 3,000 & 16 & $10^{-3}$ \\
\bottomrule
\end{tabular}
\end{table*}

\begin{table*}[t]
\centering
\small
\setlength{\tabcolsep}{7pt}
\caption{\textbf{Extended robustness results.}
All entries are per-session SIP errors in degrees.
The range test uses the model trained with
$\theta_{\max}=30^\circ$.
The re-attachment test uses the separately
trained arbitrary-mounting model.}
\label{tab:supp_robust}
\begin{tabular}{@{}llrr@{}}
\toprule
Experiment & Condition
& DIP-IMU & TotalCapture \\
\midrule
Mounting range & $\theta_{\max}=15^\circ$
& 16.7 & 13.3 \\
& $\theta_{\max}=30^\circ$
& 16.8 & 13.6 \\
& $\theta_{\max}=45^\circ$
& 17.6 & 14.7 \\
& $\theta_{\max}=60^\circ$
& 20.1 & 18.1 \\
& $\theta_{\max}=180^\circ$
& 66.6 & 65.0 \\
\midrule
Re-attachment & Unchanged, detector off
& 19.2 & 14.2 \\
& Unchanged, detector on
& 19.2 & 14.2 \\
& Changed, detector off
& 24.4 & 20.5 \\
& Changed, detector on
& 23.2 & 19.2 \\
\bottomrule
\end{tabular}
\end{table*}

\paragraph{Pose estimators.}
The default poser uses a linear projection with ReLU,
three unidirectional LSTM layers with 512 hidden units,
and a 6D rotation head for local joint rotations.
Its inputs follow TransPose's pelvis-relative
normalization.
The rotation output is supervised by
\begin{equation}
\mathcal{L}_{\mathrm{sup}}
=
\lambda_{\mathrm{rot}}\mathcal{L}_{\mathrm{rot}}
+
\lambda_{\mathrm{FK}}\mathcal{L}_{\mathrm{FK}},
\end{equation}
where the local rotation and forward-kinematic
position losses are defined as
\begin{align}
\mathcal{L}_{\mathrm{rot}}
&=\frac{1}{|\mathcal{J}_R|}
\sum_{j\in\mathcal{J}_R}
\|\hat{\mathbf{R}}_j-\mathbf{R}_j\|_F^2,\\
\mathcal{L}_{\mathrm{FK}}
&=\frac{1}{|\mathcal{J}_P|}
\sum_{j\in\mathcal{J}_P}
\|\mathrm{FK}(\hat{\mathbf{R}})_j
-\mathrm{FK}(\mathbf{R})_j\|_2^2.
\end{align}
Here, $\mathcal{J}_R$ and $\mathcal{J}_P$ denote
the joint sets used for rotation and position
supervision, respectively.
$\hat{\mathbf{R}}_j$ and $\mathbf{R}_j$ denote predicted
and ground-truth local rotation matrices.
$\mathrm{FK}$ computes root-relative joint positions
using a fixed neutral-body skeleton with identity
root rotation.
We set $\lambda_{\mathrm{rot}}=1$ and
$\lambda_{\mathrm{FK}}=10$.
Both losses are averaged over frames and batch samples.

Pretraining uses AMASS for 30,000 iterations with
batch size 64, 200-frame windows, and Adam with
a one-cycle maximum learning rate of $10^{-3}$.
Fine-tuning uses equal proportions of AMASS and
DIP-IMU subjects 1--8 for 6,000 iterations,
with batch size 32 and learning rate $10^{-4}$.
During pretraining, one or two non-pelvis sensors are
masked with probability 0.5 per window.
The invariant poser follows the same schedule and
architecture, with random mounting and common-heading
augmentation.

PIP~\cite{yi2022pip} uses released weights and its kinematic predictions
before physics-based optimization.
DynaIP~\cite{zhang2024dynaip} retains its official architecture and objective
and is retrained with the default poser's data,
iteration budgets, batch sizes, and window length.
Both networks are frozen during evaluation.


\paragraph{Re-attachment detection.}
Alongside the session sum, each sensor maintains
exponentially decayed evidence:
\begin{equation}
\mathbf{B}^{\mathrm{recent}}_{w,i}
=
2^{-\Delta t/h}\mathbf{B}^{\mathrm{recent}}_{w-1,i}
+\mathbf{A}_{w,i},
\end{equation}
where $\Delta t=1$\,s and $h=60$\,s.
After more than 60\,s since initialization or reset,
an angular discrepancy above $90^\circ$ between the
projected recent and session estimates replaces that
sensor's mounting sum with its recent sum.
Assignment evidence and other sensors remain unchanged.
This detector is enabled only in the experiments
that explicitly evaluate it.

\section{Evaluation details}
\label{sec:supp_protocol}


\paragraph{Configurations and sessions.}
Per-session evaluation concatenates a subject's recordings in
alphabetical order under a fixed configuration.
Configuration evidence persists across recordings;
the motion window, pose-network recurrent state,
and heading alignment restart at each recording.
Per-take evaluation additionally resets configuration
evidence.

For bounded mountings, the rotation axis is uniform
on the sphere and its angle follows
$U[0,\theta_{\max}]$.
Arbitrary mountings instead use uniform unit quaternions,
giving uniform sampling on $SO(3)$.

\paragraph{Metrics and aggregation.}
Following TransPose~\cite{yi2021transpose}, rotations
of ignored joints---root, ankles, feet, wrists, and
hands---are replaced by ground truth, and root
orientation is removed.
No Procrustes alignment is performed.
Mesh error uses the neutral mean-shape body model
and every fourth frame.
Pose metrics are averaged within recordings and then
across recording--configuration pairs.
Numerical differences are computed before rounding.




\paragraph{Swing and twist.}
For segment-frame mounting error
$\mathbf{E}=\mathbf{M}\hat{\mathbf{M}}^\top$,
swing measures the change of the segment's longitudinal
axis $\mathbf{u}$:
\begin{equation}
e_{\mathrm{swing}}
=
\arccos(\mathbf{u}^\top\mathbf{E}\mathbf{u}).
\end{equation}
Twist measures rotation about that axis.
When $\mathbf{u}$ is coordinate axis $k$ and
$j,l$ are the remaining axes,
\begin{equation}
e_{\mathrm{twist}}
=
\left|\operatorname{atan2}
(E_{lj}-E_{jl},E_{jj}+E_{ll})\right|.
\end{equation}
Here, $E_{lj}$ denotes the entry in row $l$ and
column $j$ of $\mathbf{E}$.
The stacked component heights are not the total
geodesic rotation error.

\paragraph{Live demo consent.}
The individual appearing in the live demo provided
written informed consent for participation and
for publication of the images in the paper.

\section{Additional results}
\label{sec:supp_results}

\paragraph{Mounting range and re-attachment.}
\Cref{tab:supp_robust} provides the full results
supporting the robustness experiments.
For the range test, the bounded model is evaluated
with uniformly sampled angles up to each indicated
$\theta_{\max}$.

The re-attachment test uses the arbitrary model
and replaces one randomly selected sensor's mounting
with a new arbitrary rotation before the middle
recording.
The detector reduces SIP error on both datasets.

\begin{figure*}[t]
\centering
\includegraphics[width=\linewidth]{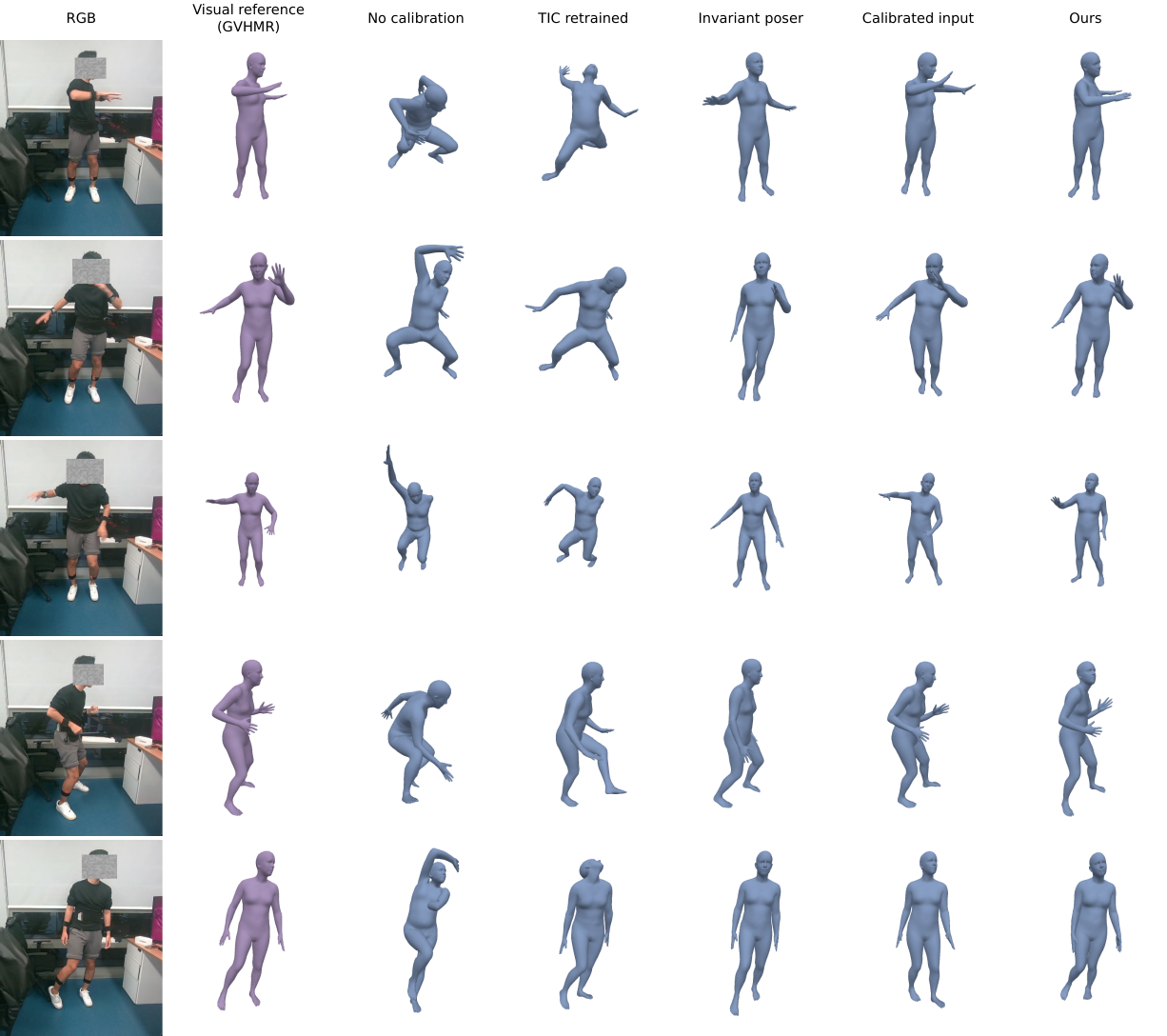}
\caption{\textbf{Qualitative comparisons on our collected IMU recordings.}
GVHMR provides the offline visual reference.
Calibrated input uses known sensor assignments and mounting
rotations derived from the initial visual pose and IMU readings,
with camera-to-IMU heading alignment estimated from the first 10\,s.}
\label{fig:supp_live}
\end{figure*}

\paragraph{Live demo}
One participant performs a variety of movements, including
walking, squats, boxing motions, and upper-body
rotations, while wearing six IMUs at the prescribed
body sites.
We record IMU measurements at 60\,Hz and RGB video
at $640\times480$ resolution and 30\,fps. We use the same \method{} and LSTM  checkpoints as in the main-paper arbitrary-mounting experiments,
without additional fine-tuning on the live recordings.
GVHMR~\cite{shen2024world} processes the video
offline to provide a visual reference.
Reference rotations are interpolated to IMU timestamps.
\Cref{tab:supp_live} reports the results for the recorded three trials, each lasting two minutes.
\Cref{fig:supp_live} shows video frames and the corresponding
SMPL reconstructions.

\begin{table*}[t]
\centering
\small
\setlength{\tabcolsep}{6pt}
\caption{\textbf{Quantitative results on our collected IMU recordings.}
Pose errors are measured against the GVHMR visual reference.
$t_{\mathrm{assign}}$ denotes the time until all six
assignments become correct and remain correct thereafter.}
\label{tab:supp_live}
\begin{tabular}{@{}lrrrrrr@{}}
\toprule
Trial
& $t_{\mathrm{assign}}$ (s)
& SIP ($^\circ$)
& Ang ($^\circ$)
& Pos (cm)
& Mesh (cm)
& Jitter (km/s$^3$) \\
\midrule
1 & 6 & 21.1 & 20.3 & 8.0 & 10.3 & 1.30 \\
2 & 7 & 22.7 & 22.8 & 9.2 & 11.8 & 1.71 \\
3 & 1 & 23.4 & 23.6 & 8.9 & 11.2 & 2.09 \\
\midrule
Mean & 4.7 & 22.4 & 22.2 & 8.7 & 11.1 & 1.70 \\
\bottomrule
\end{tabular}
\end{table*}

